\documentclass{article}

     \usepackage[preprint,nonatbib]{neurips_2019}

\usepackage[utf8]{inputenc} 
\usepackage[T1]{fontenc}    
\usepackage{hyperref}       
\usepackage{url}            
\usepackage{booktabs}       
\usepackage{amsfonts}       
\usepackage{nicefrac}       
\usepackage{microtype}      
\usepackage{graphicx}
\title{YOLO Nano: a Highly Compact You Only Look Once Convolutional Neural Network for Object Detection}

\author{\hspace{-0.5in}
  Alexander Wong$^{1,2}$, Mahmoud Famuori$^{1,2}$, Mohammad Javad Shafiee$^{1,2}$\\ \textbf{Francis Li$^2$, Brendan Chwyl$^2$, and Jonathan Chung$^2$}\\
  $^{1}$Waterloo Artificial Intelligence Institute, University of Waterloo, Waterloo, ON, Canada\\
  $^{2}$DarwinAI Corp., Waterloo, ON, Canada\\
}

\begin{document}

\maketitle

\begin{abstract}
Object detection remains an active area of research in the field of computer vision, and considerable advances and successes has been achieved in this area through the design of deep convolutional neural networks for tackling object detection.  Despite these successes, one of the biggest challenges to widespread deployment of such object detection networks on edge and mobile scenarios is the high computational and memory requirements.  As such, there has been growing research interest in the design of efficient deep neural network architectures catered for edge and mobile usage.  In this study, we introduce YOLO Nano, a highly compact deep convolutional neural network for the task of object detection.  A human-machine collaborative design strategy is leveraged to create YOLO Nano, where principled network design prototyping, based on design principles from the YOLO family of single-shot object detection network architectures, is coupled with machine-driven design exploration to create a compact network with highly customized module-level macroarchitecture and microarchitecture designs tailored for the task of embedded object detection.   The proposed YOLO Nano possesses a model size of $\sim$4.0MB ($>$15.1$\times$ and $>$8.3$\times$ smaller than Tiny YOLOv2 and Tiny YOLOv3, respectively) and requires 4.57B operations for inference (>34\% and $\sim$17\% lower than Tiny YOLOv2 and Tiny YOLOv3, respectively) while still achieving an mAP of $\sim$69.1\% on the VOC 2007 dataset ($\sim$12\% and $\sim$10.7\% higher than Tiny YOLOv2 and Tiny YOLOv3, respectively). Experiments on inference speed and power efficiency on a Jetson AGX Xavier embedded module at different power budgets further demonstrate the efficacy of YOLO Nano for embedded scenarios.
\end{abstract}
\vspace{-0.15in}
\section{Introduction}
\vspace{-0.15in}
An active area in the field of computer vision is object detection, where the goal is to not only localize objects of interest within a scene, but also assign a class label to each of these objects of interest.  Considerable recent successes in the area of object detection stems from modern advances in deep learning~\cite{lecun2015deep,krizhevsky2012imagenet}, particularly leveraging deep convolutional neural networks.  Much of the initial focus was on improving accuracy, leading to increasingly more complex object detection networks such as SSD~\cite{liu2016ssd}, R-CNN~\cite{girshick2014rich}, Mask R-CNN~\cite{MaskRCNN}, and other extended variants of these networks~\cite{huang2017speed,lin2017feature,shrivastava2016training}.  While such networks demonstrated state-of-the-art object detection performance, they were very challenging, if not impossible, to deploy on edge and mobile devices due to computational and memory constraints.  In fact, even faster variants such as Faster R-CNN~\cite{ren2015faster} have inference speeds at low single-digit frame rates when running on embedded processors. This greatly limits the widespread adoption of such networks for a wide range of applications such as unmanned aerial vehicles, video surveillance, autonomous driving where local embedded processing is required.

To address this challenge of achieving embedded object detection, there has been a growing interest in the exploration and design of highly efficient deep neural network architectures for object detection that are more well-suited for edge and mobile devices~\cite{redmon2016you,redmon2016yolo9000,YOLOv3,wu2016squeezedet,MobileNetv1,shafiee2017fast}.  A particularly interesting family of object detection networks designed around efficiency is the YOLO family of neural network architectures~\cite{redmon2016you,redmon2016yolo9000,YOLOv3}, which leverage a number of design principles to create single-shot architectures which can achieve embedded object detection performance on high-end desktop GPUs.  However, these network architectures remain too large for many edge and mobile scenarios (e.g., $\sim$240MB in the case of the YOLOv3 architecture), and their inference speeds drop considerably when running on edge and mobile processors due to computational complexity (e.g., $>$65B operations in the case of YOLOv3).  To address this issue, Redmon et al. introduced the Tiny YOLO family of network architectures, which has greatly reduced model sizes at a cost of object detection performance.

In this study, we are motivated to explore a human-machine collaborative design strategy to designing highly compact deep convolutional neural networks for the task of object detection, where principled network design prototyping is coupled with machine-driven design exploration.  More specifically, we leverage the design principles from the YOLO family of single-shot object detection network architectures within this human-machine collaborative design strategy to create YOLO Nano, a highly compact network with highly customized module-level macroarchitecture and microarchitecture designs tailored for the task of embedded object detection.

\vspace{-0.15in}
\section{Methods}
\label{method}
\vspace{-0.15in}
In this study, we introduce YOLO Nano, a highly compact deep convolutional neural network for embedded object detection designed using a human-machine collaborative design strategy~\cite{AttoNets}.  The human-machine collaborative design strategy for designing YOLO Nano comprises of two main design stages: i) principled network design prototyping, and ii) machine-driven design exploration.
\vspace{-0.15in}
\subsection{Principled network design prototyping}
\vspace{-0.12in}
The first design stage in creating YOLO Nano is a principled network design prototyping stage, where we create an initial network design prototype (denoted as $\varphi$), based on human-driven design principles to guide the machine-driven design exploration stage.  More specifically, we construct an initial network design prototype based on the design principles of the YOLO family of single-shot architecture~\cite{redmon2016you,redmon2016yolo9000,YOLOv3}.  A standout characteristic of the YOLO family of network architectures is that, unlike region proposal-based networks which rely on the construction of a regional proposal network to generate proposals for where objects lie in the scene followed by classification on the generated proposals, they instead leverage a single network architecture to process the input image and generate the output results.  As such, all object detection predictions for a single image are made in a single forward pass, compared to hundreds to thousands of passes that need to be performed to get the final results for region proposal-based networks.  This makes the YOLO family of network architectures significantly faster to run, and thus better suited for embedded object detection.

The initial design prototype used in this study draws inspiration from the YOLO family of network architectures and is comprised of a stack of feature representation modules, with shortcut connections between the modules as with~\cite{YOLOv3}.  Also, as with~\cite{YOLOv3}, the feature representation modules are configured in a way, similar to feature pyramid networks~\cite{featurepyramidnetworks}, such that it is capable of representing features at three different scales.  These feature representation modules are followed by several convolutional layers, with output being a three-dimensional tensor that encodes bounding box, objectness, and class predictions for three different scales.  As a result, this initial design prototype architecture design allows for efficient multi-scale object detection.

The actual macroarchitecture and microarchitecture designs of the individual  modules and layers in the final YOLO Nano network architecture, as well as the number of network modules, are left for the machine-driven design exploration stage to determine automatically given data as well as human-specified design requirements and constraints designed specifically around edge and mobile scenarios with limited computational and memory capabilities.

\vspace{-0.12in}
\subsection{Machine-driven design exploration}
\vspace{-0.12in}
Using the initial network design prototype ($\varphi$), data, as well as human-specified design requirements catered to edge and mobile usage as a guide, a machine-driven design exploration stage is then leveraged to determine the module-level macroarchitecture and microarchitecture designs for the proposed YOLO Nano network architecture.  More specifically, machine-driven design exploration is achieved in this study in the form of generative synthesis~\cite{Wong2018}, which is capable of determining the optimal macroarchitecture and microarchitecture designs of the final network architecture within the human-specified requirements and constraints.  The overall goal of generative synthesis is to learn generative machines that can generate deep neural networks that meet design requirements and constraints, and can be described as follows. This is formulated within the concept of generative synthesis as a constrained optimization problem for determining a generator $\mathcal{G}$ that, given a set of seeds $S$, can generate networks $\left\{N_s|s \in S\right\}$  maximizing a universal performance function $\mathcal{U}$ (e.g.,~\cite{Wong2018_Netscore}) while satisfying requirements and constraints defined via an indicator function $1_r(\cdot)$:
\vspace{-0.05in}
\begin{equation}
\mathcal{G}  = \max_{\mathcal{G}}~\mathcal{U}(\mathcal{G}(s))~~\textrm{subject~to}~~1_r(\mathcal{G}(s))=1,~~\forall s \in S.
\label{optimization}
\vspace{-0.05in}
\end{equation}
Since it is computationally intractable to solve for the globally optimal solution in the constrained optimization problem posed in Eq.~\ref{optimization} given the enormity of the feasible region, we instead solve for an approximate solution $\hat \mathcal{G}$ via iterative optimization, where the initial solution $\hat \mathcal{G}_0$ is guided by $\varphi$, $\mathcal{U}$, and $1_r(\cdot)$, and progressively updated such that each successive approximate solution $\hat \mathcal{G}_k$ achieving a higher $\mathcal{U}$ than previous approximate solutions (i.e., $\hat \mathcal{G}_1$, $\ldots$, $\hat \mathcal{G}_{k-1}$, etc.) while still constrained by $1_r(\cdot)$.  The final approximate solution $\hat \mathcal{G}$ is then used to create the proposed YOLO Nano network.

To guide the generative synthesis process towards learning generative machines that generate object detection networks for edge and mobile scenarios that are not only highly efficient and compact but also provide strong object detection performance, one of the key steps is to configure the indicator function $1_r(\cdot)$ to enforce the appropriate design requirements and constraints.  In this study, the indicator function $1_r(\cdot)$ was set up such that: i) mean average precision (mAP) $\geq$ 65\% on VOC 2007, ii) computational cost $\leq$ 5B operations, and iii) 8-bit weight precision.  The computational cost constraint is set such that the computational cost of the resulting YOLO Nano network is below that of Tiny YOLOv3~\cite{YOLOv3}, one of the most popular compact networks for embedded object detection.

\vspace{-0.15in}
\section{YOLO Nano Architectural Design}
\label{design}
\begin{figure}[t]
\begin{center}
			\includegraphics[width = 0.9\linewidth]{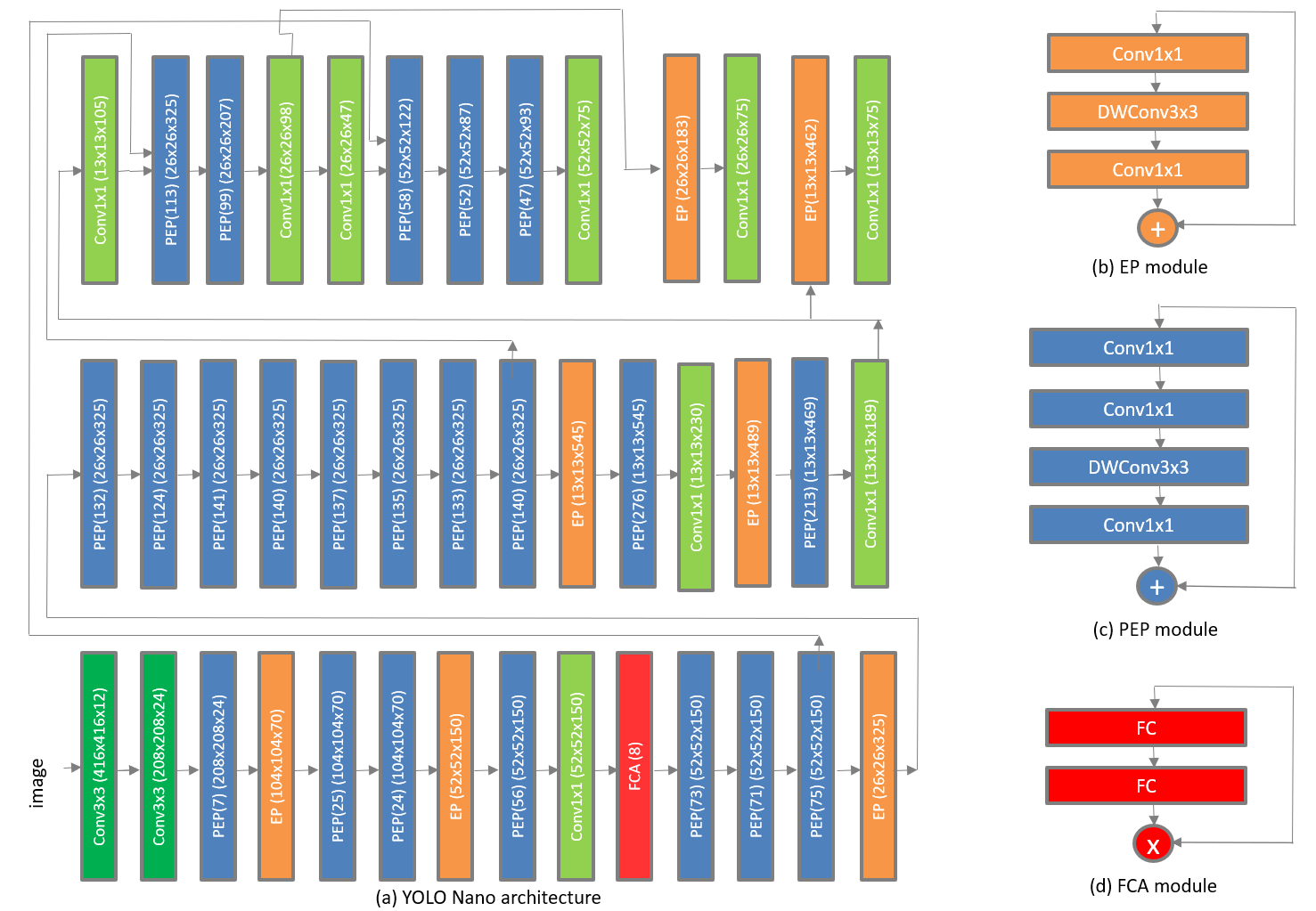}
			\vspace{-0.1in}
		\caption{YOLO Nano network architecture. Note that PEP(\textit{x}) indicates \textit{x} channels in the first projection layer of a residual PEP module, and FCA(\textit{x}) indicates reduction ratio of \textit{x}}\vspace{-0.2in}
		\label{fig:yolonano}
\end{center}
\vspace{-0.1in}
\end{figure}

\vspace{-0.15in}
The network architecture of the proposed YOLO Nano network for embedded object detection is shown in Figure~\ref{fig:yolonano}, with several interesting observations worth discussing below.
\vspace{-0.1in}

\vspace{-0.05in}

\subsection{Residual Projection-Expansion-Projection Macroarchitecture}
\vspace{-0.1in}
The first notable observation about the YOLO Nano network architecture that differs significantly from the YOLO family of networks is that it is comprised of modules with unique residual projection-expansion-projection (PEP) macroarchitectures, in addition to expansion-projection (EP) macroarchitectures like those found in~\cite{MobileNetv2,mnas,fairnas}.  The residual PEP macroarchitecture consists of: i) a projection layer with 1$\times$1 convolutions that projects output channels into an output tensor with lower dimensionality ii) an expansion layer with 1$\times$1 convolutions, that expands the number of channels to a higher dimensionality, iii) a depth-wise convolution layer that performs spatial convolutions with a different filter on each of the the individual output channels from the expansion layer, and iv) a projection layer with 1$\times$1 convolutions that projects output channels into an output tensor with lower dimensionality.  The use of residual PEP macroarchitectures enables significant reductions in the architectural and computational complexity while preserving model expressiveness.
\vspace{-0.1in}
\subsection{Fully-connected Attention Macroarchitecture}
\vspace{-0.15in}
The second notable observation about the YOLO Nano network architecture is the strategic introduction of light-weight fully-connected attention (FCA) within the network by the machine-driven design exploration process, which is in contrast to fixed module-level introduction in other design exploration methods~\cite{mnas}.  As with~\cite{squeezeexcite}, the FCA macroarchitecture consists of two fully-connected layers that learn the dynamic, non-linear inter-dependencies between channels and produces modulation weights for re-weight the channels via channel-wise multiplication. The use of FCA facilitates for dynamic feature recalibration based on global information to pay more attention to informative features, thus enabling better utilization of available network capacity.  This in turn allows for a strong balance between reduced architectural and computational complexity and model expressiveness.

\vspace{-0.1in}
\subsection{Macroarchitecture and Microarchitecture Heterogeneity}
\vspace{-0.15in}

The third notable observation about the YOLO Nano network architecture is that there is high heterogeneity in terms of not only macroarchitectures (a diverse mix of PEP modules, EP modules, FCA, as well as individual 3$\times$3 and 1$\times$1 convolution layers), but also in terms of the microarchitectures of the individual feature representation modules and layers, with each module or layer in the network having unique microarchitectures.  The benefit of having high microarchitecture heterogeneity in the YOLO Nano network architecture is that it enables each component of the network architecture to be uniquely tailored to achieve a very strong balance between architectural and computational complexity and model expressiveness. This architectural diversity in YOLO Nano also demonstrates the advantage of leveraging a machine-driven design exploration strategy as flexible as generative synthesis as it would be impossible for a human designer, or other design exploration methods such as \cite{mnas,fairnas} to customize a network architecture to this level of architectural granularity.

\vspace{-0.3 cm}
\section{Experimental Results and Discussion}
\vspace{-0.35 cm}
To study the efficacy of YOLO Nano for  embedded object detection, we examine its model size, object detection accuracy, and computational cost on the PASCAL VOC datasets.  For comparison purposes, the Tiny YOLOv2 network~\cite{redmon2016yolo9000} and the Tiny YOLOv3 network~\cite{YOLOv3} were used as a baseline references given that they are amongst the most popular compact deep neural networks for embedded object detection given their small model sizes and low computational complexities.  The VOC2007/2012 datasets consist of natural images that have been annotated with 20 different types of objects.  The deep neural networks were trained using the VOC2007/2012 training datasets, and the mean average precision (mAP) was computed on the VOC2007 test dataset to evaluate the object detection accuracy of the deep neural networks, as is standard practice in research literature.

Table~\ref{Tab:res} shows the model sizes and the object detection accuracies of the proposed YOLO Nano network as well as Tiny YOLOv2 and Tiny YOLOv3.  First, it was observed that the model size of YOLO Nano was 4.0MB, which is $>$15.1$\times$ and $>$8.3$\times$ smaller than Tiny YOLOv2 and Tiny YOLOv3, respectively, which is very important for edge and mobile scenarios given the memory constraints.  Second, YOLO Nano, despite being much smaller in model size, achieved an mAP of 69.1\% on the VOC 2007 test dataset, which is $\sim$12\% and $\sim$10.7\% higher than that of Tiny YOLOv2 and Tiny YOLOv3, respectively.  Third, YOLO Nano requires just 4.57 billion operations to perform inference, which is >34\% lower than Tiny YOLOv2 and $\sim$17\% lower than Tiny YOLOv3.
\vspace{-0.05in}

\begin{table}[ht]
	\begin{center}
		\small
		\caption{Object detection accuracy results of tested compact networks on VOC 2007 test set. Input size is 416$\times$416 for all tested networks. Best results are highlighted in \textbf{bold}.}
\vspace{-0.05in}
		\label{Tab:res}
		\begin{tabular}{|c|c||c|c|c|c|c|}
			\hline
			Model  & Model  & mAP & computational cost \\
			Name & size  & (VOC 2007) & (ops) \\\hline \hline
			Tiny YOLOv2~\cite{redmon2016yolo9000} & 60.5MB &57.1\% & 6.97B\\
			Tiny YOLOv3~\cite{YOLOv3} & 33.4MB &58.4\% & 5.52B\\	
			YOLO Nano & \textbf{4.0MB} & \textbf{69.1\%} & \textbf{4.57B} \\\hline
		\end{tabular}
	\end{center}
\vspace{-0.15in}
\end{table}

Finally, to investigate the real-world performance of YOLO Nano within an embedded scenario, we evaluated the inference speed and power efficiency of YOLO Nano running on a Jetson AGX Xavier embedded module at different power budgets.  At 15W and 30W power budgets, YOLO Nano achieved inference speeds of $\sim$26.9 FPS and $\sim$48.2 FPS, respectively, resulting in power efficiencies of $\sim$1.97 images/sec/watt and $\sim$1.61 images/sec/watt, respectively.  These experimental results show that the proposed YOLO Nano network, created through a human-machine collaborative design strategy, provides a strong balance between accuracy, size, and computational complexity that makes it well suited for embedded object detection for edge and mobile scenarios.
{\small
	\bibliographystyle{abbrv}
	\bibliography{ccn_style}
}
\end{document}